# UAV Marketplace Simulation Tool for BVLOS Operations

## Demonstration Track

Kıvanç Şerefoğlu
Özyegin Üniversity
Istanbul, Turkiye
kivanc.serefoglu@ozyegin.edu.tr

Önder Gürcan
NORCE
Norwegian Research Center AS
Kristiansand, Norway
ongu@norceresearch.no

Reyhan Aydoğan
Özyegin Üniversity and Delft
University of Technology
Istanbul and Delft, Turkiye and NL
reyhan.aydogan@ozyegin.edu.tr

## ABSTRACT

We present a simulation tool for evaluating team formation in autonomous multi-UAV (Unmanned Aerial Vehicle) missions that operate Beyond Visual Line of Sight (BVLOS). The tool models UAV collaboration and mission execution in dynamic and adversarial conditions, where Byzantine UAVs attempt to disrupt operations. Our tool allows researchers to integrate and compare various team formation strategies in a controlled environment with configurable mission parameters and adversarial behaviors. The log of each simulation run is stored in a structured way along with performance metrics so that statistical analysis could be done straightforwardly. The tool is versatile for testing and improving UAV coordination strategies in real-world applications.

## 1 INTRODUCTION

Autonomous unmanned aerial vehicles (UAVs) are becoming integral to diverse applications, including precision agriculture, environmental monitoring, military surveillance, and search and rescue missions [3, 12]. As UAV technology advances, ensuring their dependability is critical, particularly in high-stakes missions where failures can result in operational disruptions, financial losses, or risks to human lives. Reliability, safety, and robustness are fundamental to optimize their effectiveness in various missions, strengthening public trust, and facilitating greater adoption in the commercial and public sectors [15]. Fully autonomous multi-UAV operations require highly reliable systems to guarantee success in complex real-world applications. The dependability in such operations encompasses the robustness [8], resilience [11], and fault tolerance [13] of autonomous systems. As UAV technology evolves towards complete autonomy [14], it becomes crucial to ensure that these aerial vehicles can perform their tasks under varying conditions without human intervention.

One particularly promising application lies in UAV marketplaces, where UAVs from different owners, who may not necessarily trust each other, collaborate to form teams or swarms for Beyond Visual Line of Sight (BVLOS) missions [2, 5, 7]. These UAVs are selected based on their reputation, ensuring reliable and efficient collaboration. During such missions, UAVs autonomously adhere to predefined protocols, logging critical events, behaviors, and any observed misbehavior. At the end of the mission, the UAVs evaluate their teammates and operators based on these observations, enabling quick and efficient reputation updates. Hence, we present a simulation tool [1] to evaluate UAV collaboration strategies and team formation mechanisms in BVLOS missions [2]. Drawing inspiration from the Fully Autonomous UAV (FAU) teamwork case study [2], the tool provides a lightweight framework for realistic mission modeling without using blockchain. It incorporates configurable mission parameters and adversarial behaviors to simulate dynamic and complex environments. The tool facilitates the comparison of team formation strategies with performance metrics conveniently exported in CSV format for in-depth analysis.

## 2 UMS-SIM: UAV MARKETPLACE SIMULATOR

We consider a UAV Marketplace System (UMS) containing several marketplaces $U = \{A_1, A_2, ..., A_i\}$ where $i$ is the number of the marketplace. Each marketplace $A_i$ is denoted as $A_i = \{C_i, B_i\}$ where $C_i$ is the set of cooperative UAVs and $B_i$ is the set of Byzantine UAVs, and we assume that $A_i$ is fixed during missions, i.e. no UAV is added or removed. The degree of cooperativeness and malicious behavior (e.g., False Feedback Attacks [4, 10], Collusion Attacks [9, 10]) may vary for the agents. Collaboration among Byzantine UAVs can be disabled if required. Marketplaces are classified by difficulty levels based on the proportion of Byzantine UAVs, with higher percentages presenting greater challenges and requiring robust strategies to mitigate adversarial behaviors. Additionally, marketplaces can include diverse UAV types with varying features such as speed, sensor range, and battery capacity. Each mission in the simulation is defined using a flexible template that outlines key components, allowing customization of mission objectives and parameters. UMS can contain several missions $M = \{M_1, M_2, ..., M_i\}$ where $i$ is the number of the mission. As shown in Figure 1, there is a pool of marketplaces, each containing different UAVs. The operator can create new mission templates and marketplaces, as well as adding UAVs to these marketplaces. Once team formation strategies are implemented, the operator selects a marketplace and mission to proceed with. The UMS then forms a team from the chosen marketplace based on the selected team formation strategy. Upon completion of the mission, the UAVs provide feedback to the UMS, which updates their reputation scores accordingly based on the submitted reports.

---

[1] Gitlab, https://gitlab.com/scop-framework/model-library/security-and-trust/scop-lib-st-rtf
[2] https://www.youtube.com/watch?v=h0pt29Cx08s

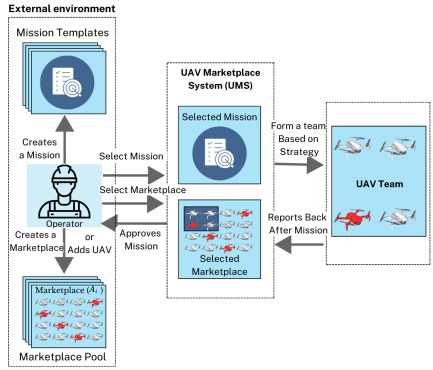

Figure 1: UMS

UMS-Sim is implemented using the SCOP framework[3], a Java-based framework that enables the modeling and execution of agent-based systems in dynamic and adversarial environments. This system provides a flexible and configurable environment for analyzing how UAVs collaborate, form teams, and complete missions under various conditions. The simulation tool organizes necessary data, such as marketplaces and missions, using JSON files for efficient and structured data management. Log files and CSV files are generated after mission completion, providing a record of the simulation's results and performance metrics, such as finish time, UAVs' reputation score change, and which UAVs are selected. UMS-Sim currently has three different team formation strategies implemented: MDBR [1], EigenTrust [6], and Random Selection. Each strategy is executed sequentially within an episode cycle. During each cycle, a different strategy is applied in each episode under the same environmental conditions. Once all implemented strategies have been executed, the cycle restarts. The environment remains unchanged if the configuration specifies constancy; otherwise, it adjusts after each cycle. This process continues until the predetermined number of episodes is complete. At the end of each episode, the simulation tool extracts data into CSV files, organized by the team formation strategy used. These files include key details such as the marketplace name, the percentage of Byzantine UAVs, the UAVs involved in the mission, their scores, and mission-specific information like completion time and whether UAVs behaved adversarially. In addition to CSV files, each UAV in the simulation has its own log file, where all its actions during the mission are recorded. These logs provide a clear record of every decision and interaction, making it easy to analyze individual UAV behavior.

Our simulation tool includes an LLM chat panel that allows users to interact with agents, inquire about UAV behavior, and understand mission outcomes. Additionally, communication with the UMS is enabled, making it possible to ask broader questions, such as which team formation strategy performs better than others.

[3]SCOP framework, https://scop-framework.netlify.app

## 3 CASE STUDY

We present a use case focusing on a scenario in which UAVs must locate a stationary human target in a forest fire before the fire reaches it (see Figure 2a). Each mission $M_i$ is denoted as: $M_i = \{n, d, t, p, s, h, c\}$ where $n$ represents the total number of UAVs deployed in the mission, $d$ is the forest density, $t$ denotes the fire spread time (tick), representing the rate at which the fire spreads across the area, $p$ indicates the fire start position, $s$ defines the area size as row × column dimensions and $h$ specifies the position of the stationary human target that must be located. Finally, $c$ indicates whether Byzantine UAV collaboration is enabled ($c$=1) or disabled ($c$=0). The search strategy used in these missions is the *sweeping* strategy. Agents employ sweeping moves from west to east step by step, and move forward when they complete all cells on the same line. In our case study, we compare the three previously mentioned strategies: MDBR, EigenTrust, and Random Selection as seen in Figure 2b.

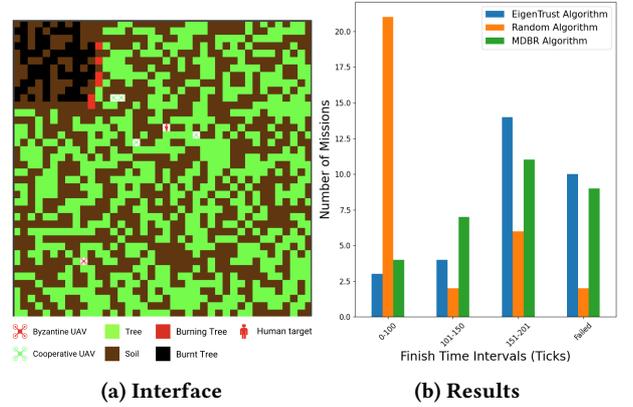

(a) Interface  (b) Results

Figure 2: Simulation

Our case study includes scenarios where Byzantine UAVs collaborate by assigning high-reputation scores to themselves while giving low scores to other UAVs. Furthermore, these Byzantine UAVs send false coordination information, claiming the presence of a target within 10 cells near their location [16]. This misleads other UAVs, causing the team to waste time searching for the real target. When cooperative UAVs receive a message, they travel to the specified coordinates to verify the presence of the target. If the target is not found at the given location, the sender of the message is flagged as an unreliable source. For example, as clearly illustrated in Figure 2b, the EigenTrust and MDBR algorithms are manipulated by Byzantine UAVs. Therefore, in this marketplace, the Random Selection algorithm is the most effective compared to the others.

## 4 CONCLUSION

The UMS-Sim tool has been developed to be a user-friendly and easy to use tool to compare team formation strategies for multi-UAV BVLOS missions. It enables researchers to evaluate UAV collaboration under dynamic and adversarial conditions, addressing the challenges posed by Byzantine UAVs.